%% file: main.tex
\documentclass[letterpaper]{article} 
\usepackage{aaai25}  
\usepackage{times}  
\usepackage{helvet}  
\usepackage{courier}  
\usepackage[hyphens]{url}  
\usepackage{graphicx} 
\usepackage{natbib}  
\usepackage{caption} 
\usepackage{booktabs}
\usepackage{amsmath}
\usepackage{algorithm}
\usepackage{algorithmic}
\usepackage{multirow}
\usepackage{bm}
\usepackage{subcaption}
\usepackage{diagbox}

\usepackage{enumitem}

\newcommand{\stexttt}[1]{#1}

\usepackage{newfloat}
\usepackage{listings}
\DeclareCaptionStyle{ruled}{labelfont=normalfont,labelsep=colon,strut=off} 
\floatstyle{ruled}
\newfloat{listing}{tb}{lst}{}
\floatname{listing}{Listing}
\title{Merge and Bound: Direct Manipulations on Weights for Class Incremental Learning}

\author{$\text{Taehoon Kim}^{1}$ \quad\quad\quad \; $\text{Donghwan Jang}^{2}$ \quad\quad\quad \; $\text{Bohyung Han}^{3,4}$\\
}

\affiliations{
     $\text{School of Informatics, University of Edinburgh}^{1}$ \\ 
$\text{Siebel School of Computing and Data Science, UIUC}^{2}$ \\
$\text{ECE}^{3}$ \& $\text{IPAI}^{4}$, Seoul National University\\
{\tt\small \{kthone, jh01120, bhhan\}@snu.ac.kr}
}

\usepackage{bibentry}

\begin{document}

\maketitle

\input{section/0_abstract}

\input{section/1_introduction}
\input{section/2_related}

\input{section/3_method}

\input{section/4_exp}
\input{section/5_conclusion}

\paragraph{Acknowledgements}
This work was partly supported by Samsung Advanced Institute of Technology (SAIT).
\bibliography{aaai25}
\clearpage

\end{document}

%% file: section/0_abstract.tex

\begin{abstract}
We present a novel training approach, named Merge-and-Bound (M\&B) for Class Incremental Learning (CIL), which directly manipulates model weights in the parameter space for optimization.
Our algorithm involves two types of weight merging: \textit{inter-task weight merging} and \textit{intra-task weight merging}.
Inter-task weight merging unifies previous models by averaging the weights of models from all previous stages.
On the other hand, intra-task weight merging facilitates the learning of current task by combining the model parameters within current stage.
For reliable weight merging, we also propose a bounded update technique that aims to optimize the target model with minimal cumulative updates and preserve knowledge from previous tasks; this strategy reveals that it is possible to effectively obtain new models near old ones, reducing catastrophic forgetting.
M\&B is seamlessly integrated into existing CIL methods without modifying architecture components or revising learning objectives.
We extensively evaluate our algorithm on standard CIL benchmarks and demonstrate superior performance compared to state-of-the-art methods.

\end{abstract}

%% file: section/1_introduction.tex

\section{Introduction}
Despite the remarkable achievements of recent deep neural networks (DNNs)~\cite{radford2021learning, ho2020denoising, brown2020language}, training under continually shifting data distributions encounters a significant challenge called \textit{catastrophic forgetting}---significant performance degradation on previously learned data. 
Since the real-world environments, in which DNNs are deployed, dynamically change over time, addressing this issue becomes critical for enhancing the efficiency and applicability of DNNs.

Class Incremental Learning (CIL)~\cite{douillard2020podnet, hou2019learning, rebuffi2017icarl, simon2021learning, kang2022class} is a kind of continual learning, which deals with a sequential influx of tasks, typically composed of disjoint class sets.
Given the constraint of permitting to store only a few or no examples from previous tasks during training for new ones, catastrophic forgetting becomes a primary obstacle for CIL.
Prior approaches have attempted to tackle this issue through methods including knowledge distillation~\cite{hinton2015distilling,romero2014fitnets,zagoruyko2016paying,kang2022class}, architecture expansion~\cite{rusu2016progressive,yoon2018lifelong,liu2021adaptive,yan2021dynamically,abati2020conditional}, or parameter regularization~\cite{aljundi2018memory,kirkpatrick2017overcoming,zenke2017continual}. 
However, these methods have inherent limitations, such as dependency on data from previous tasks for distilling knowledges of previous tasks, the need for additional network components, and poor performance, which hinder their wide-range applications.

We propose a novel training approach for CIL, referred to as Merge-and-Bound (M\&B), which can be easily integrated into existing CIL methods without any modifications on architecture or loss function.
Motivated by recent studies~\cite{wortsman2022model, rame2022diverse, wortsman2022robust, izmailov2018averaging}, demonstrating the benefits of weight averaging in aggregating the capabilities of multiple models, we introduce two types of weight merging techniques tailored for complex task dynamics of CIL---inter-task weight merging and intra-task weight merging---which respectively enhance the stability and plasticity of a CIL model.
In order to preserve all the knowledge acquired up to the current task, \textit{inter-task weight merging} averages the parameters of the models learned from all incremental stages in an online fashion to form a base model.
The base model serves as an initialization point for a subsequent task. 
On the other hand, \textit{intra-task weight merging} improves the model's adaptivity to new tasks.
This technique improves the model's generalization ability for each new task by averaging multiple checkpoints along the training trajectory within the current task.

Furthermore, we incorporate \textit{bounded model update}~\cite{tianTrainableProjectedGradient2023, gouk2020distance} for training models in each task, which constrains weight updates within the neighborhood of the initial model in each incremental stage.
By preventing model parameters from deviating excessively from the base model, this strategy preserves knowledge gained from previous tasks. 
When this strategy is employed in conjunction with weight averaging, we anticipate a more reliable weight merging results by reducing the variation of merged models and leading them to stay within the same basin of the objective function with respect to the previous tasks. 
Throughout the integration of these techniques, we maintain a balance between stability, preserving knowledge from previous tasks, and plasticity, adapting to new tasks in CIL scenarios.

The contributions of this paper are summarized as follows:

\begin{itemize}
	\item[$\bullet$] We propose two weight merging techniques specialized to CIL, inter- and intra-task weight merging. 
	These techniques enhance the model's stability and plasticity by averaging model parameters across tasks (inter-task) and within each task (intra-task), respectively. \vspace{0.1cm}
	\item[$\bullet$]  We introduce a bounded model update strategy that constrains the total amount of model updates within each task. 
	By enforcing the new models to remain close to the old ones, our approach alleviates the catastrophic forgetting of previously acquired knowledge and encourage reliable weight merging.%
\vspace{0.1cm}
		 \item[$\bullet$] Our algorithm, which can be conveniently integrated into existing CIL methods, substantially and consistently improves performance on multiple benchmarks with marginal extra computational complexity. 
\end{itemize}

\vspace{1mm}

%% file: section/2_related.tex
\section{Related Works}
\label{sec:related}

\subsection{Class Incremental Learning (CIL)}
CIL is a challenging problem that aims to learn a model with the number of classes increasing stage-by-stage without forgetting the previously learned classes.
We organize CIL methods into five groups based on their main strategies: parameter regularization, architecture expansion, bias correction, knowledge distillation, and rehearsal methods. 

Parameter regularization methods~\cite{aljundi2018memory,kirkpatrick2017overcoming,zenke2017continual} measure the importance of network parameters and adjust their flexibility to mitigate catastrophic forgetting. 
However, these methods suffer from unsatisfactory generalization performance in CIL scenarios~\cite{van2019three,hsu2018re}.
Architecture expansion methods~\cite{rusu2016progressive,yoon2018lifelong,liu2021adaptive,yan2021dynamically,abati2020conditional} dynamically expand the network capacity to handle incoming tasks by adding new neurons or layers. However, they introduce computational burdens due to additional network components.
Bias correction methods~\cite{hou2019learning,wu2019large} address the bias towards new classes caused by the class imbalances in CIL by introducing scale and shift parameters or matching the scale of weight vectors.
Knowledge distillation methods~\cite{hinton2015distilling,romero2014fitnets,zagoruyko2016paying,kang2022class} encourage models to preserve previous task knowledge by mimicking the representations of old models. 
Several approaches match output distributions or attention maps to preserve important information.
Rehearsal-based methods~\cite{rebuffi2017icarl,ostapenko2019learning,shin2017continual} store representative examples or employ generative models to mitigate forgetting. Examples include maintaining class centroids~\cite{rebuffi2017icarl} or using generative adversarial networks~\cite{goodfellow2014generative,liu2020generative,odena2017conditional} to generate synthetic examples.

In contrast, we propose a novel weight manipulation approach which is compatible and easy to integrate with the existing CIL methods, without requiring any changes to the network architectures or loss functions. 

\begin{figure*}[t!]
    \centering
    \begin{tabular}{cc}
        \includegraphics[width=0.45\textwidth]{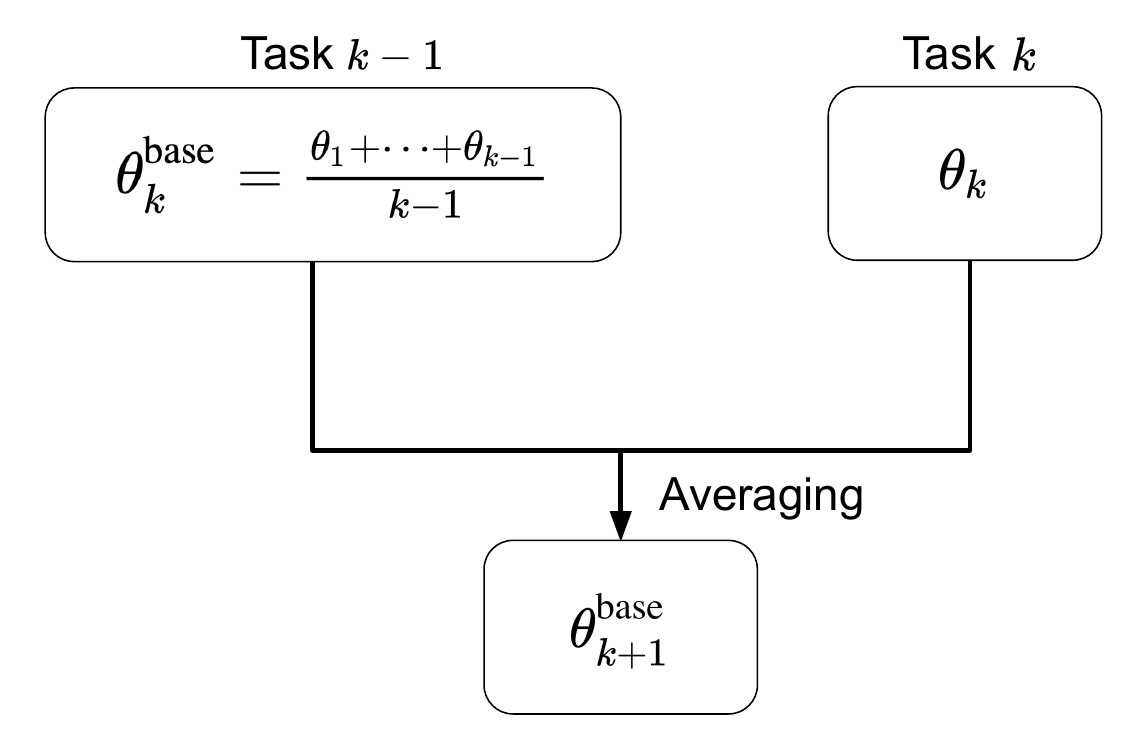} \quad\quad &
        \includegraphics[width=0.45\textwidth]{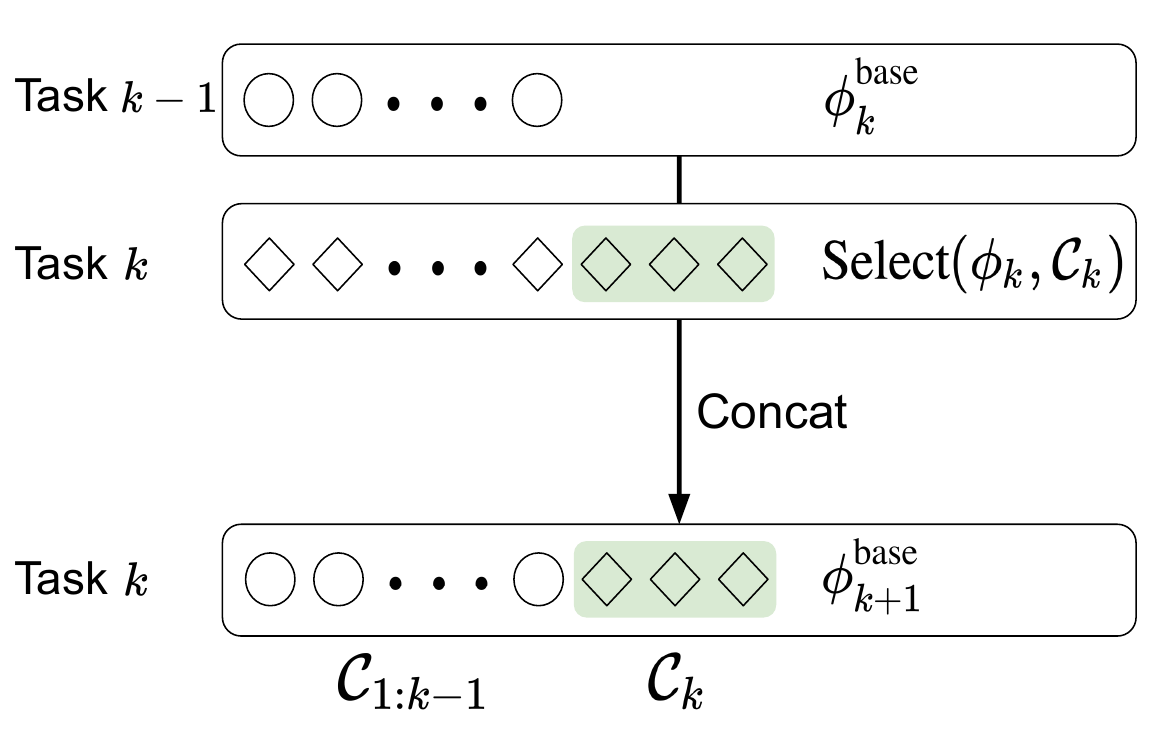} \\
        (a) Feature extractor	$f_{\theta_{k+1}^{\text{base}}}$ & (b) Classifier $g_{\phi_{k+1}^{\text{base}}}$
    \end{tabular}
\caption{
    Description of inter-task weight merging: Upon the completion of the \( {k}^{\text{th}} \) incremental stage, we establish the base model \( M^{\text{base}}_{k+1}(\cdot) \), which will serve as the \textit{initialization point} for the \( (k+1)^{\text{st}} \) stage.
    The model comprises a feature extractor \( f_{\theta_{k+1}^{\text{base}}}(\cdot) \) and a classifier \( g_{\phi_{k+1}^{\text{base}}}(\cdot) \), and they are constructed by the following procedures.
    (a) To construct the base feature extractor \( f_{\theta_{k+1}^{\text{base}}}(\cdot) \), we set \( {\theta_{k+1}^{\text{base}}} \) to the moving average of all the previous feature extractor weights, \( \theta_1, \theta_2, \cdots, \theta_{k} \), which is easily computed with \( \theta_k^{\text{base}} \) and \( \theta_k \) in a recursive manner following Equation~\eqref{eq:feature_extractor_base}.
    (b) For learning the classifier \( g_{\phi_{k+1}^{\text{base}}}(\cdot) \), we concatenate the weights of the current base classifier \(\phi_{k}^{\text{base}} \) with the weights of the current classifier \( {\phi_{k}} \) associated with the class set in the current task \( \mathcal{C}_{k} \).
}
    \label{fig:base}
    \vspace{-4mm}
\end{figure*}

\subsection{Robust Transfer Learning}
Robust transfer learning aims to adapt a pre-trained model to a new domain or task without losing its generalization ability. 
This is similar to continual learning, which seeks to prevent catastrophic forgetting of previous knowledge while learning new tasks sequentially. 

A common approach to robust transfer learning is to freeze or constrain the weights of the pre-trained model~\cite{kumarFineTuningCanDistort2022,leeSurgicalFineTuningImproves2022} while fine-tuning. 
This prevents feature distortion from adaptation and preserves the original knowledge. 
Another line of work explores the idea of weight averaging or interpolation to enhance the performance of fine-tuned models~\cite{neyshaburWhatBeingTransferred2020,wortsman2022model,rameModelRatatouilleRecycling2023}. 
However, these methods primarily concentrate on a single downstream task, neglecting the sequential domain shifts that are inherent in continual learning.

%% file: section/3_method.tex

\section{Proposed Approaches}
\label{sec:method}

\subsection{Problem Formulation}
CIL is a learning framework to handle a sequence of tasks, $\mathcal{T}_{1:K} = \{ T_1, \cdots, T_K \}$.
Each task $T_k$ consists of a labeled dataset $\mathcal{D}_k$ whose label set, $\mathcal{C}_k$, is disjoint to the ones defined in the past, \textit{i.e.}, $(\mathcal{C}_1 \cup \cdots \cup \mathcal{C}_{k-1}) \cap \mathcal{C}_k = \emptyset$.
At the $k^{\text{th}}$ incremental stage, the current model $M_k(\cdot)$ 
is trained on integrated dataset $\mathcal{D}'_k = \mathcal{D}_k \cup \mathcal{B}_{k-1}$, where $\mathcal{B}_{k-1}$ is a set of representative exemplars that belong to all the previously learned classes.
The performance of a CIL algorithm is evaluated using a test set comprising test data from all the stages.

\subsection{Motivation}
Inspired by recent studies in transfer learning~\cite{wortsman2022model, rame2022diverse, wortsman2022robust, izmailov2018averaging} that combine the weights of multiple models, we propose two unique model averaging strategies for CIL in the presence of complex dynamics in data distribution: inter-task weight merging and intra-task weight merging.
Although both techniques concentrate on integrating the competency of multiple models, they have distinct objectives. 
Inter-task weight merging aims to consolidate knowledge from all the previously learned tasks and construct a comprehensive model by computing the moving average of the previous model parameters.
Meanwhile, intra-task weight merging boosts the model's generalization capability on new task by merging the multiple models along the training trajectory of a current task.

To boost the effectiveness and reliability of both weight merging techniques, we introduce bounded model update scheme.
Recent studies on model averaging~\cite{neyshaburWhatBeingTransferred2020,wortsman2022model,rameModelRatatouilleRecycling2023} highlight that weight averaging techniques are effective for the models associated with the same loss basin, leading to identify a flat minimum within the basin.
However, in the scenarios that data from previous tasks are scarce during incremental learning stages, training on the current task with a large number of training examples results in the convergence to different local minima from the one associated with earlier tasks.
To tackle the issue, we impose a constraint on the magnitude of model updates and maintain the robustness of trained models to diverse tasks over many incremental stages.
Through the bounded model update strategy for training on new tasks, trained models are less prone to stray from the loss basins of previous tasks. 
This approach stabilizes our weight merging techniques by reducing the variance of trained models across incremental stages and within each of an incremental task, facilitating the achievement of consistent performance.

In the rest of this section, we describe the details of each technical components, inter-task weight averaging, intra-task weight averaging, and bounded model update.


 \begin{figure*}[t!]
    \centering
    \begin{subfigure}[b]{0.46\textwidth}
        \centering
        \includegraphics[width=\textwidth]{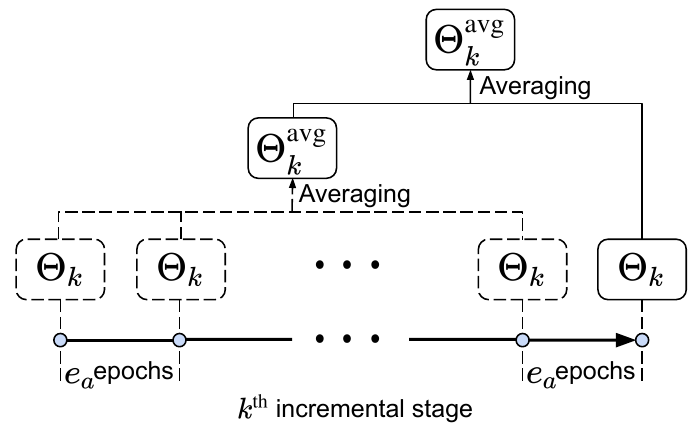} 
        \caption{Intra-task weight Merging}
        \label{fig:weight_averaging}
    \end{subfigure}
    \quad\quad \quad 
    \begin{subfigure}[b]{0.405\textwidth}
        \centering
        \includegraphics[width=\textwidth]{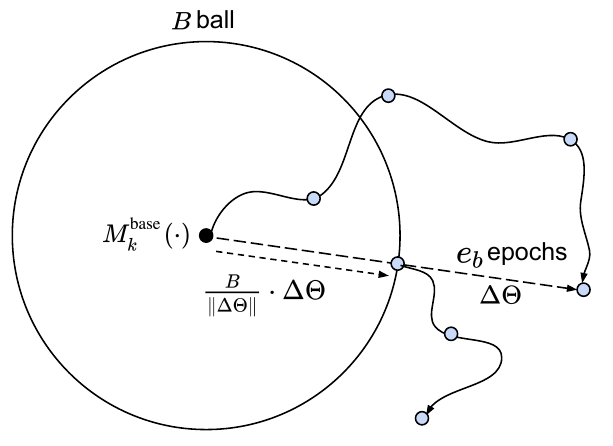} 
        \caption{Bounded Updates}
        \label{fig:bounded_updates}
    \end{subfigure}
    
\caption{
    (a) Illustration of the intra-task weight merging:
We introduce an intra-task weight merging by the moving average of weights in multiple models along the training trajectories, as described in Equation~\eqref{eq:intra_task_averaging}. 
Intra-task weight merged model is utilized for inference and for computing the next stage base model \(M_k^{\text{base}}(\cdot)\).
(b) Illustration of the bounded update technique:
We constrain the weight updates around the base model denoted by \(M_k^{\text{base}}(\cdot)\).
This strategy is designed to preserve the knowledge in the base model but search for the unexplored space during optimization.
}
    \label{fig:intra}
    \vspace{-4mm}
\end{figure*}

\subsection{Inter-task weight merging}
We merge the models from all the preceding tasks to the base model $M_k^{\text{base}}(\cdot)$, for consolidating the previously acquired knowledge.
The base model is composed of a feature extractor $f_{\theta_k^{\text{base}}}(\cdot)$ and a classifier $g_{\phi_k^{\text{base}}}(\cdot)$, collectively parameterized by $\Theta_k^{\text{base}} = \{\theta_k^{\text{base}}, \phi_k^{\text{base}}\}$.

Upon the completion of the $k^{\text{th}}$ incremental stage, we combine the current feature extractor $f_{\theta_k}(\cdot)$ and classifier $g_{\phi_k}(\cdot)$ into the base model, $M_k^{\text{base}}(\cdot)$ at the corresponding stage, as illustrated in Figure~\ref{fig:base}. 
This process creates the new base model for the next stage, denoted by $M_{k+1}^{\text{base}}(\cdot)$.
The new feature extractor is given by the moving average of all models as follows:
\begin{equation}
\theta_{k+1}^{\text{base}} = \frac{k-1}{k} \cdot \theta_{k}^{\text{base}} + \frac{1}{k} \cdot \theta_k.
\label{eq:feature_extractor_base}
\end{equation}
To obtain the classifier of the new base model at the $(k+1)^\text{st}$ stage, we concatenate the current base model classifier $g_{\phi_k^{\text{base}}}(\cdot)$ defined for the classes in $\mathcal{C}_{k-1}$  with the weights corresponding to the classes in $\mathcal{C}_k$ of the current classifier, $g_{\phi_k}(\cdot)$, which is expressed as
\begin{equation}
\phi_{k+1}^{\text{base}} = \text{Concat}(\phi_{k}^{\text{base}}, \text{Select}(\phi_k, \mathcal{C}_k)),
\end{equation}
where $\text{Select}(\phi_k, \mathcal{C}_k)$ extracts the weights corresponding to the classes in $\mathcal{C}_k$ of the current classifier $g_{\phi_k}(\cdot)$.
At the ${(k+1)}^{\text{st}}$ incremental stage, we set the initial model as the base model, $M_{k+1}^{\text{base}}(\cdot)$, which encapsulates all the knowledge learned up to the current stage.

\subsection{Intra-task weight merging}

The primary goal of intra-task weight merging is to enhance generalization ability on the current task by averaging the multiple checkpoints along training trajectories.
In the $k^{\text{th}}$ incremental stage, an intra-task merged model, $M^{\text{avg}}_k(\cdot)$, parameterized by $\Theta_k^{\text{avg}}$ is updated at every $e_a$  epochs as
%
\begin{equation}
\Theta_k^{\text{avg}} \leftarrow \frac{n \cdot \Theta_k^{\text{avg}}  + \Theta_k}{n+1},
\label{eq:intra_task_averaging}
\end{equation}
where $n$ denotes the number of models involved in the merging.
Once the stage is completed, the final model, $M_k(\cdot)$, is replaced by the intra-task merged model, $M^{\text{avg}}_k(\cdot)$, \textit{i.e.}, $\Theta_k\leftarrow\Theta_k^\text{avg}$, for inference and the computation of the base model, $M_{k+1}^{\text{base}}(\cdot)$.

For the models equipped with Batch Normalization (BN)~\cite{ioffe2015batch}, an extra post-training data forwarding is required to estimate new mean and variance of the activations after model averaging~\cite{garipov2018loss,izmailov2018averaging}.
To this end, instead of executing an additional forward pass after resetting the running statistics as in SWA~\cite{izmailov2018averaging}, we conduct the forward pass on top of the running statistics of the current model $M_k(\cdot)$ estimated before model merging, which incurs slight updates of the BN statistics.
This strategy alleviates the bias towards the current task caused by sample deficiency of the classes introduced in the previous tasks.

\begin{table*}[t]
    \caption{CIL performance (\%) on CIFAR-100.
    The proposed training technique (M\&B) consistently improves performance when plugged into the existing methods.
    Note that we run 3 experiments with 3 different orders and report the average result. 
    Models with $*$ denotes the report of reproduced results.    
    The bold-faced numbers indicate the best performance.}
    \label{tab:cifar}
    \vspace{-3mm}
    \begin{center}
        \scalebox{0.85}{
            \setlength\tabcolsep{5pt}
            \renewcommand{\arraystretch}{1}
            \hspace{-4mm}
            \begin{tabular}{ccccc}
                \toprule
                &\multicolumn{4}{c}{CIFAR-100}  \\ 
                Number of tasks & 5 &  10 &  25 &  50  \\ 
                \midrule
                iCaRL~\cite{rebuffi2017icarl} & 58.08 & 53.78 & 50.60 & 44.20  \\
                BiC~\cite{wu2019large} & 56.86 & 53.21 & 48.96 & 47.09  \\
                Mnemonics~\cite{liu2020mnemonics} &  63.34 &  62.28 &  60.96 & \hspace{0.34cm}-  \\ 
                GeoDL{*}~\cite{simon2021learning} & 65.34 &  63.61 &  60.21 &  52.28  \\
                UCIR~\cite{hou2019learning} & 64.01 & 61.22 &  57.57 & 49.30 \\  \midrule
                PODNet{*}~\cite{douillard2020podnet} &64.83$\pm$0.62 &62.75$\pm$0.74 & 60.73$\pm$0.62& 58.37$\pm$0.83  \\
                PODNet~\cite{douillard2020podnet} + M\&B &\textbf{67.69}$\pm$0.50 &\textbf{66.49}$\pm$0.50 & \textbf{64.93}$\pm$0.42& \textbf{63.29}$\pm$0.37  \\  \midrule
                AFC{*}~\cite{kang2022class} &66.11$\pm$0.60 &64.77$\pm$0.74 & 63.68$\pm$0.74& 61.94$\pm$0.60\\
                AFC~\cite{kang2022class} + M\&B &\textbf{67.96}$\pm$0.70 &\textbf{67.10}$\pm$0.68 & \textbf{66.12}$\pm$0.67& \textbf{65.38}$\pm$0.41 \\ \midrule
                FOSTER{*}~\cite{wang2022foster} &72.23$\pm$0.51 &69.12$\pm$0.67 & 65.45$\pm$0.90& 59.60$\pm$0.85\\
                FOSTER~\cite{wang2022foster} + M\&B &\textbf{73.03}$\pm$0.61 &\textbf{70.58}$\pm$0.64 & \textbf{66.79}$\pm$0.84& \textbf{62.47}$\pm$0.97 \\
                \bottomrule	
            \end{tabular} 
        }
    \end{center}
       \vspace{-4mm}
\end{table*}

\subsection{Bounded Model Update}
\label{sec:bd}


We enforces a constraint for model update in CIL, which bounds the magnitude of weight updates from the base model at every $e_b$ epochs as 
\begin{equation}
    \Delta\Theta \leftarrow \begin{cases}
    B \cdot \frac{\Delta\Theta}{\| \Delta\Theta \|}, & \text{if } \| \Delta\Theta \| > B \\
    \Delta\Theta, & \text{otherwise}
\end{cases},
\end{equation}
where $\Delta\Theta$ denotes the displacement of the current model from the base model, $M_k^{\text{base}}(\cdot)$, and $B$ indicates the limit of the total model update magnitude.
Note that the proposed bounded model update is performed within an incremental stage jointly with intra-task weight merging.
Given that the base model is presumed to hold the knowledge of previous tasks, this strategy aims to directly prevent model updates in the current task from diverging from the base model in the weight space, thereby preventing the loss of previously learned information.
By combining bounded updates with inter-task and intra-task weight merging, this approach is expected to foster an reliable ensemble effect.

%% file: section/4_exp.tex
\section{Experiments}

\label{sec:exp}

\begin{table*}[t!]
    \caption{CIL performance (\%) on ImageNet-100/1000.
    M\&B demonstrates significant performance gains when integrated into existing methods, even in the large-scale benchmarks for CIL.}
    \label{tab:img}
    \vspace{-3mm}
    \begin{center}
        \scalebox{0.85}{
            \setlength\tabcolsep{8pt}
            \renewcommand{\arraystretch}{1}
            \hspace{-2mm}
            \begin{tabular}{cccc|cc}
                \toprule
                &\multicolumn{3}{c}{ImageNet-100} & \multicolumn{2}{c}{ImageNet-1000} \\ 
                Number of tasks & 5 &  10 &  25  & 5 &  10  \\ 
                \midrule
                iCaRL~\cite{rebuffi2017icarl} &  65.56 & 60.90 & 54.56 & 51.36 & 46.72 \\
                BiC~\cite{wu2019large} & 68.97 & 65.14 & 59.65 & 45.72 & 44.31 \\
                Mnemonics~\cite{liu2020mnemonics} & 72.58 &  71.37 &  69.74 & 64.54 & 63.01 \\ 
                GeoDL{*}~\cite{simon2021learning} & 73.87 &  73.55 &  71.72 & 65.23 & 64.46\\
                UCIR~\cite{hou2019learning} & 71.04 & 70.71 &  62.94 & 64.34 & 61.18 \\  \midrule
                PODNet{*}~\cite{douillard2020podnet} &74.06 &71.51 & 67.31 & 68.18 & 65.58\\
                PODNet~\cite{douillard2020podnet} + M\&B &\textbf{75.98} &\textbf{74.08} & \textbf{70.70}&\textbf{69.53} & \textbf{67.76}  \\  \midrule
                AFC{*}~\cite{kang2022class} & 76.91 &75.26 & 73.65&68.06&66.39 \\
                AFC~\cite{kang2022class} + M\&B &\textbf{77.05} &\textbf{76.35} & \textbf{74.35} &\textbf{70.28} & \textbf{69.51} \\ \midrule
                FOSTER{*}~\cite{wang2022foster} & 80.22 &78.15 & 71.74&--&--\\
                FOSTER~\cite{wang2022foster} + M\&B &\textbf{80.59} &\textbf{79.20} & \textbf{73.24} &--&-- \\ 
                \bottomrule	
            \end{tabular} 
        }
    \end{center}
       \vspace{-4mm}
\end{table*}

\subsection{Datasets and Evaluation Protocol}
We conduct experiments on CIFAR-100~\cite{cifar09} and ImageNet-100/1000~\cite{ILSVRC15}. 
CIFAR-100 contains 50,000 and 10,000 training and validation images, respectively, in 100 classes.
Following the standard protocols, we evaluate all the compared algorithms using three different class orders on CIFAR-100 and the class order specified in~\cite{douillard2020podnet} on ImageNet-100 and ImageNet-1000.

We incorporate Merge-and-Bound (M\&B) into various state-of-the-art CIL algorithms based on knowledge distillation (PODNet~\cite{douillard2020podnet}, AFC~\cite{kang2022class}), architecture expansion (FOSTER~\cite{wang2022foster}), and virtual class augmentation (IL2A~\cite{zhu2021class}).
Note that IL2A is employed to compare with non-exemplar-based methods.

As in the previous works~\cite{douillard2020podnet,hou2019learning,liu2021adaptive}, we first train the model using a half of the entire classes at the initial stage and split the remaining classes into 5/10/25/50 stages on CIFAR-100, 5/10/25 stages on ImageNet-100, and 5/10 stages on ImageNet-1000 to simulate CIL scenarios.
We test models on all the seen classes at each incremental stage, and report the \textit{average incremental accuracy}~\cite{rebuffi2017icarl,douillard2020podnet,hou2019learning}---average accuracy over all incremental stages.

\subsection{Implementation Details}
\label{sec:imp}
As our approach is a plug-in method, we follow the implementation settings of the existing methods~\cite{douillard2020podnet, kang2022class,wang2022foster} in principle. 
The memory budget size is set to 20 per class unless specified otherwise.
Detailed description about implementation details and hyperparameters are discussed in supplementary document.

\subsection{Results on CIFAR-100 and ImageNet-100/1000}

\subsubsection{CIFAR-100}
Table~\ref{tab:cifar} illustrates that the proposed algorithm, denoted by M\&B, enhances the performance of the baseline models in all CIL scenarios, with notable margins.
The performance gains of M\&B appear modest in FOSTER~\cite{wang2022foster}.
This is because of the design choice of FOSTER~\cite{wang2022foster}, which exploits enlarged model capacity and ensemble effects from multiple models.
Such strategies tend to reduce the unique benefits of the proposed approach, especially when the number of incremental stages is small.
However, the performance of the proposed method stands out again when the number of incremental stages becomes 50, where the benefits of FOSTER~\cite{wang2022foster} diminish abruptly due to lack of model capacity.

\subsubsection{ImageNet-100/1000}
Table~\ref{tab:img} shows the results on large-scale benchmarks, ImageNet-100/1000.
While the proposed method consistently boosts the baseline algorithms, Table~\ref{tab:img} clearly illustrates that the performance gains are particularly remarkable for ImageNet-1000, which is the most challenging benchmark for CIL.
Additionally, across all datasets, the proposed algorithm demonstrates enhanced performance gains with an increasing number of tasks. 
This characteristic is highly desirable for CIL in practical scenarios, which have no restrictions or information on the number of incoming tasks.
Note that we were unable to reproduce the ImageNet-1000 experiments for FOSTER~\cite{wang2022foster} since the training configuration was not released publicly.
 
\subsection{Ablation Studies}
\label{sec:abl}
We perform various ablation studies to validate the effectiveness of the proposed training technique.
All the experiments are performed on CIFAR-100 with 50 incremental stages unless specified otherwise.
To measure the stability and plasticity, we compute the average new accuracy, which is the average accuracy of new classes over incremental stages, and the forgetting metric~\cite{lee2019overcoming}, which is the average of the performance degradation for each class.

\begin{table}[t!]
    \caption{Component analysis of our algorithm for inter-task weight merging, intra-task weight merging, and bounded model update. 
    }
	\label{tab:abl}
	   \vspace{-3mm}
	   	\begin{center}
	\setlength\tabcolsep{2pt}	
	\scalebox{0.85}{
\begin{tabular}{lccc}
\toprule
 & Forgetting $\downarrow$ & Avg. new acc. $\uparrow$ & Overall acc.$\uparrow$ \\ 
\midrule
\midrule
(a) M\&B & 15.38  &  59.35  & \textbf{65.38}   \\ \midrule
(b) \stexttt{w/o inter-task} & 21.77 & 62.74 & 61.81  \\ 
(c) \stexttt{w/o intra-task} & \textbf{13.10} & 51.60 & 65.08  \\ 
(d) \stexttt{w/o bounded} & 18.72 & \textbf{64.14} & 64.21  \\ 
 
\bottomrule
\end{tabular}
}
\end{center}
   \vspace{-5mm}
\end{table}

\begin{table}[t!]
    \caption{CIL performance (\%) with a limited memory budget on CIFAR-100: 1 exemplar memory per class for PODNet~\cite{douillard2020podnet} and AFC~\cite{kang2022class}, and no memory for IL2A~\cite{zhu2021class}.
	}
    \label{tab:mem1}
   \vspace{-3mm}
    \begin{center}
        \scalebox{0.85}{
            \setlength\tabcolsep{2pt}
            \renewcommand{\arraystretch}{1}
            \hspace{-3mm}
            \begin{tabular}{cccccc}
                \toprule
                &\multicolumn{4}{c}{CIFAR-100}  \\ 
                Number of tasks & 5 &  10 &  25 &  50  \\ 
                \midrule        
                PODNet{*}~\cite{douillard2020podnet} &43.70 &34.19 & 26.58& 14.78  \\
                PODNet~\cite{douillard2020podnet} + M\&B &\textbf{52.27}&\textbf{45.63} & \textbf{36.81}& \textbf{20.84} \\  \midrule
                AFC{*}~\cite{kang2022class} &49.82 &42.78& 35.51& 23.59 \\
                AFC~\cite{kang2022class} + M\&B&\textbf{53.41} &\textbf{48.92} & \textbf{46.08}& \textbf{35.25} \\  \midrule
                IL2A{*}~\cite{zhu2021class} & 65.70 &58.14& 54.40& 20.42  \\
                IL2A~\cite{zhu2021class} + M\&B  &\textbf{67.46} &\textbf{61.80} & \textbf{58.25}& \textbf{43.54}  \\
                \bottomrule	
            \end{tabular} 
        }
    \end{center}
       \vspace{-3mm}
\end{table}

\subsubsection{Variations of the proposed method}
Table~\ref{tab:abl} presents the results obtained from different combinations of components in M\&B. 
Each component contributes to the overall improvements, albeit in different ways. 
Comparing M\&B to \stexttt{w/o inter-task} and \stexttt{w/o bounded}, we observe a decrease in forgetting, indicating that inter-task weight merging and bounded updates effectively help retain previous knowledge. 
However, the low average new accuracy in \stexttt{w/o intra-task} suggests that integrating bounded updates solely with inter-task weight merging leads to significant degradation in adaptation. 
On the other hand, the increase in average new accuracy in ours compared to \stexttt{w/o intra-task} indicates that the inclusion of intra-task weight merging, in combination with bounded updates, helps mitigate this problem. 
Thus, the combination of these components enhances the performance of our method in retaining previous knowledge while facilitating adaptation to new tasks.

\subsubsection{Results on limited memory budgets}
We evaluate the performance of M\&B with exemplar-based methods such as PODNet~\cite{douillard2020podnet} and AFC~\cite{kang2022class} when only one exemplar is available.
We also verify the effectiveness of M\&B when it is combined with non-exemplar-based method, IL2A~\cite{zhu2021class}.
As shown in Table~\ref{tab:mem1}, our algorithm significantly outperforms existing methods when operating under limited memory budgets by exploiting model merging techniques and constraining the amount of model updates. 
This property is desirable conceptually because CIL may have a large number of stages and even a small number of exemplars may be difficult to hold in practice.

 \begin{figure*}[t!]
    \centering
    \begin{subfigure}[b]{0.45\textwidth}
        \centering
        \includegraphics[width=\textwidth]{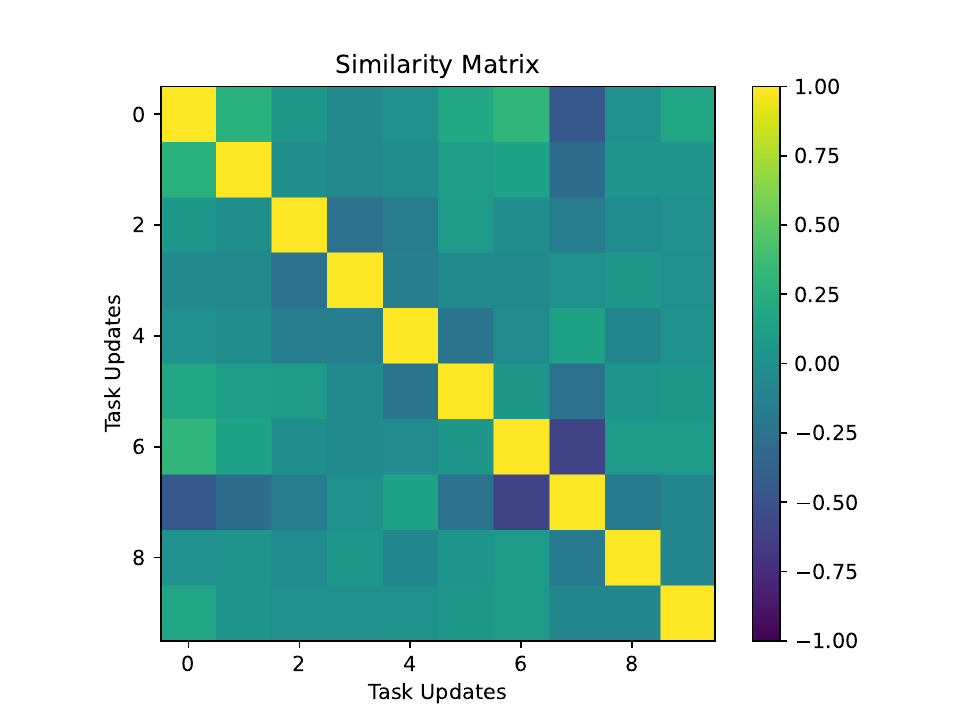} 
      
        \caption{Task similarity of AFC~\cite{kang2022class}}
        \label{fig:afcmine}
    \end{subfigure}
   \hfill
    \begin{subfigure}[b]{0.45\textwidth}
        \centering
        \includegraphics[width=\textwidth]{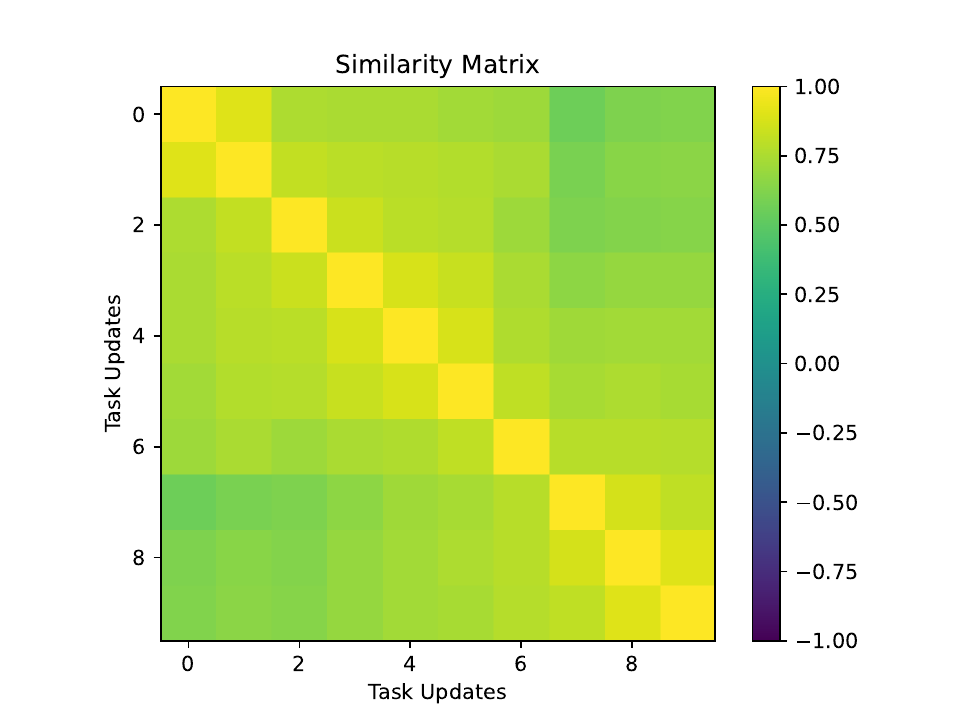} 

        \caption{Task similarity of AFC~\cite{kang2022class} + M\&B}
        \label{fig:afc}
    \end{subfigure}   
    \vspace{-2mm}
\caption{
We measure the cosine similarities between all pairs of the model update vectors occurred in each stage. 
The model update vectors become positively correlated when M\&B is incorporated.
}
    \label{fig:task_sim}
    \vspace{-5mm}
\end{figure*}

\begin{figure*}[t!]
    \centering
    \begin{subfigure}[b]{0.45\textwidth}
        \centering
        \includegraphics[width=\textwidth]{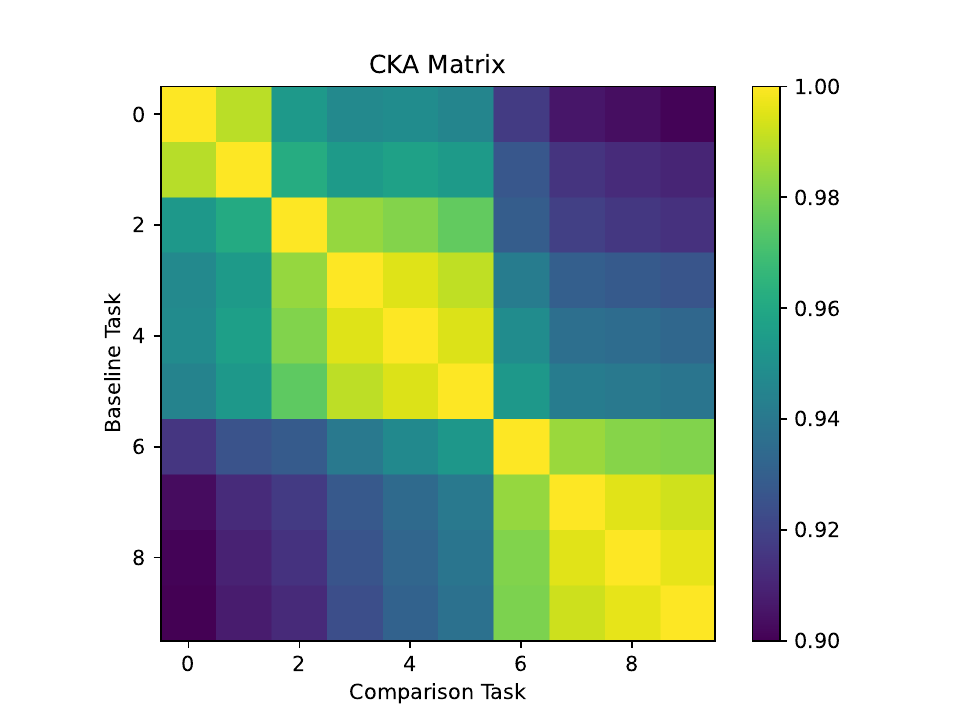} 
        \caption{CKA of AFC~\cite{kang2022class}}
        \label{fig:afc_cka}
    \end{subfigure}
   \hfill
    \begin{subfigure}[b]{0.45\textwidth}
        \centering
        \includegraphics[width=\textwidth]{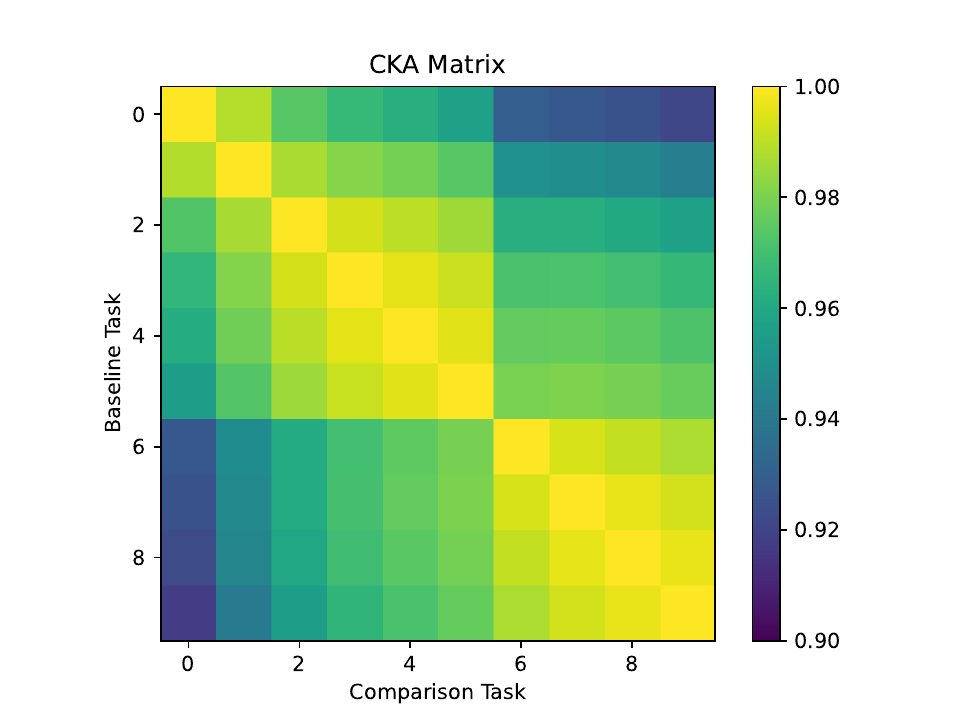} 
        \caption{CKA of AFC~\cite{kang2022class} + M\&B}
        \label{fig:afcmine_cka}
    \end{subfigure}
    \vspace{-2mm}
\caption{
CKA between models after training individual incremental stages. 
We visualize the similarity between pairs of models obtained from two different tasks---baseline task and comparison task---by measuring CKA of the representations of test examples of all classes learned up to baseline task extracted from the two models.
}
    \label{fig:cka}
    \vspace{-5mm}
\end{figure*}

\begin{table}[t!]
    \caption{Analysis of inter-task weight merging. 
    EMA denotes exponential moving average, where the numbers in parentheses indicate the smoothing factor; a higher smoothing factor assigns more weights to the most recent tasks. 
    }
    \label{tab:inter}
      \vspace{-3mm}
    \begin{center}
        \scalebox{0.85}{
            \setlength\tabcolsep{3pt}
            \renewcommand{\arraystretch}{1}
            \hspace{-3mm}
            \begin{tabular}{ccccc}
               \toprule
               &\multicolumn{4}{c}{Averaging factor} \\
                  & \shortstack[l]{Avg (Ours)} &  \shortstack[l]{EMA  (0.9)} &  \shortstack[l]{EMA  (0.5)} & \shortstack[l]{EMA  (0.1)} \\ 
                \midrule        
                Forgetting $\downarrow$ & \textbf{15.38} & 19.12 & 18.28& 17.41   \\
                Avg. new acc. $\uparrow$ &59.35 &61.34 & \textbf{61.74} & 59.10 \\  \midrule
                Overall acc. $\uparrow$ & \textbf{65.38} &64.09 & 64.10& 64.01   \\               
                \bottomrule	
            \end{tabular} 
        }
    \end{center}
       \vspace{-5mm}
\end{table}

\subsubsection{Variations in inter-task weight merging}
Table~\ref{tab:inter} shows the results from different strategies of inter-task weight merging.
Our moving average technique provides the same solution with the offline averaging, which outperforms Exponential Moving Averaging (EMA) with various smoothing factors.
The EMA methods favor recent models and lead to forgetting the previous knowledge. 

\subsubsection{Variations in intra-task weight merging}
We explore the characteristics of intra-task weight merging by varying the weight averaging periods and the BN statistics computation strategies.
According to Table~\ref{tab:intra}, the proposed intra-task weight merging technique is robust to the changes of averaging period but the results are affected by the methods to compute the BN statistics.
Due to the distinct characteristics of CIL that it should perform well on the previous classes, resetting the running statistics~(R) gives severe performance degradation since the computed running statistics will be highly biased towards current tasks in computation procedure after the reset.
Also, not forwarding the additional data path~(NC) shows the degraded performance since there is discrepancy between the running statistics and intra-task merged model since the statistics for the merged model are not computed.

\subsubsection{Variations in bounded model updates}
We conducted experiments by varying the frequency and the allowed size of the bounded model updates.
As illustrated in Table~\ref{tab:bound}, the overall accuracy is robust to the changes in the bounding frequency while applying the bounded model update results in clear advantage.
For the bounding threshold, we observe that larger threshold results in larger adaptivity since it allows model to change in large magnitude for the new task.
\vspace{-3mm}

 \begin{table}[t!]
    \caption{Results by varying the weight averaging periods and the BN statistics computation methods in intra-class weight merging. 
    R means resetting the running statistics and NC indicates no change of the BN statistics after the weight merging.
	{$\dagger$} denotes our choice for reporting the main results.
	}
    \label{tab:intra}
    \vspace{-3mm}
    \begin{center}
        \scalebox{0.85}{
            \setlength\tabcolsep{4pt}
            \renewcommand{\arraystretch}{1}
            \hspace{-3mm}
            \begin{tabular}{ccccc|ccc}
               \toprule
               &\multicolumn{4}{c}{Weight averaging period} & \multicolumn{3}{c}{BatchNorm}\\
                  & $\text{1}^\dagger$ & 5 &  10 &  15 & Ours &  R &  NC \\ 
                \midrule        
                Forgetting $\downarrow$ &\textbf{15.38} & 16.12& 15.59 & 15.58 & \textbf{15.38}  & 36.67&25.60 \\
                Avg. new acc. $\uparrow$ &59.35 & \textbf{59.41} & 58.50 & 58.95 & 59.35 & 39.75 & \textbf{60.47} \\  \midrule
                Overall acc. $\uparrow$ &65.38 &64.85 & 65.19 & \textbf{65.51} & \textbf{65.38} & 25.60 & 64.57  \\
               
                \bottomrule	
            \end{tabular} 
        }
    \end{center}
       \vspace{-3mm}
\end{table}

\begin{table}[t!]
    \caption{Results by varying the bounding period and the size of the bound. 
	}
    \label{tab:bound}
   \vspace{-3mm}
    \begin{center}
        \scalebox{0.85}{
            \setlength\tabcolsep{5pt}
            \renewcommand{\arraystretch}{1}
            \hspace{-3mm}
            \begin{tabular}{cccc|cccc}
                \toprule
               &\multicolumn{3}{c}{Bounding period} & \multicolumn{3}{c}{Bounding threshold}\\
                    &  5 &  10 & $\text{15}^{\dagger}$  &5 & $\text{10}^\dagger$  & 15    \\ 
                \midrule        
                Forgetting $\downarrow$  & 15.89 & 16.45 &\textbf{15.38}& 16.53& \textbf{15.38} &16.33  \\
                Avg. new acc. $\uparrow$  & 57.92 & 58.12 &\textbf{59.35}  & 58.66 & 59.35& \textbf{59.66}\\  \midrule
                Overall acc. $\uparrow$ & 64.81 & 64.73  &\textbf{65.38}  &64.50 & \textbf{65.38}& 64.99   \\
               
                \bottomrule	
            \end{tabular} 
        }
    \end{center}
       \vspace{-5mm}
\end{table}

\subsubsection{Effects of M\&B on weights updates}
To better comprehend the impact of our method, M\&B, within the weight space, we conducted an analysis of the cosine similarity among task updates. 
Task updates are defined by the weight changes from the beginning to the end of each task. 
As depicted in Figure~\ref{fig:task_sim}, the absence of M\&B results in task updates that are largely independent, and in some cases, exhibit a negative correlation. 
Conversely, the integration of M\&B leads to a significant positive correlation among task updates. 
These aligned task updates enable stable integration of different tasks and reduce the catastrophic forgetting by preventing the learning of new tasks from disrupting previous task updates.

\subsubsection{Similarity between representations}
To assess the impact of the proposed method on feature representations, we measure the similarity of representations.
To this end, we extract the feature representations of the test examples from the final layer of the models trained on different tasks, {\it e.g.}, baseline task and comparison task using Centered Kernel Alignment (CKA)~\cite{cortes2012algorithms, kornblith2019similarity} using test data of all classes learned up to the baseline task.
Figure~\ref{fig:cka} clearly illustrates that M\&B enhances feature similarities across the models trained in different incremental stages, which implies that M\&B alleviates catastrophic forgetting at representation level.

\subsubsection{Computational cost}

We provide precise training time overhead induced by our method. 
Based on a single NVIDIA RTX-8000 GPU with a ResNet-32 backbone, we observe that the inter-  and intra-task weight merging only take 0.003 seconds each, whereas the bounded model update operation requires 0.011 seconds. 
Given that these operations only occur intermittently, {\it e.g.}, per task or every several epochs, the additional training time is marginal.
Note that one epoch of additional forwarding is required for computing BN statistics per task.
Because M\&B is a training technique, it incurs no extra overhead for inference.
These results demonstrates that the proposed method only requires negligible computational cost and shows the great performance gains.

%% file: section/5_conclusion.tex

\section{Conclusion}
\label{sec:conclusion}
We introduce an unique Class Incremental Learning (CIL) approach that incorporates weight ensemble techniques to counteract catastrophic forgetting.
By redefining CIL as a sequential transfer learning process, innovative inter-task and intra-task weight averaging, as well as bounded update techniques were introduced. 
The proposed strategy is seamlessly integrated into existing CIL strategies without necessitating changes to the network architecture or loss function.
Additionally, it exceeded the performance of contemporary methods on numerous benchmarks, specifically in low memory budget scenarios. 
This paper contributes significantly to CIL research by elucidating how model weights can be effectively employed to maintain and transfer knowledge across tasks. 

%% file: aaai25.bib
@inproceedings{radford2021learning,
  title={Learning transferable visual models from natural language supervision},
  author={Radford, Alec and Kim, Jong Wook and Hallacy, Chris and Ramesh, Aditya and Goh, Gabriel and Agarwal, Sandhini and Sastry, Girish and Askell, Amanda and Mishkin, Pamela and Clark, Jack and others},
  booktitle={ICML},
  year={2021},
}

@inproceedings{ho2020denoising,
  title={Denoising diffusion probabilistic models},
  author={Ho, Jonathan and Jain, Ajay and Abbeel, Pieter},
  booktitle={NeurIPS},
  year={2020}
}

@inproceedings{brown2020language,
  title={Language models are few-shot learners},
  author={Brown, Tom and Mann, Benjamin and Ryder, Nick and Subbiah, Melanie and Kaplan, Jared D and Dhariwal, Prafulla and Neelakantan, Arvind and Shyam, Pranav and Sastry, Girish and Askell, Amanda and others},
  booktitle={NeurIPS},
  year={2020}
}

@inproceedings{lee2019overcoming,
  title={Overcoming catastrophic forgetting with unlabeled data in the wild},
  author={Lee, Kibok and Lee, Kimin and Shin, Jinwoo and Lee, Honglak},
  booktitle={Proceedings of the IEEE/CVF International Conference on Computer Vision},
  pages={312--321},
  year={2019}
}

@inproceedings{yoon2018lifelong,
  title="{Lifelong Learning with Dynamically Expandable Networks}",
  author={Yoon, Jaehong and Yang, Eunho and Lee, Jeongtae and Hwang, Sung Ju},
  booktitle={ICLR},
  year={2018}
}

@article{rusu2016progressive,
  title="{Progressive Neural Networks}",
  author={Rusu, Andrei A and Rabinowitz, Neil C and Desjardins, Guillaume and Soyer, Hubert and Kirkpatrick, James and Kavukcuoglu, Koray and Pascanu, Razvan and Hadsell, Raia},
   journal={arXiv preprint arXiv:1606.04671},
  year={2016}
}

@inproceedings{yan2021dynamically,
  title="{DER: Dynamically Expandable Representation for Class Incremental Learning}",
  author={Yan, Shipeng and Xie, Jiangwei and He, Xuming},
  booktitle={CVPR},
  year={2021}
}

@inproceedings{simon2021learning,
  title="{On Learning the Geodesic Path for Incremental Learning}",
  author={Simon, Christian and Koniusz, Piotr and Harandi, Mehrtash},
  booktitle={CVPR},
  year={2021}
}

@inproceedings{kang2022class,
  title={Class-incremental learning by knowledge distillation with adaptive feature consolidation},
  author={Kang, Minsoo and Park, Jaeyoo and Han, Bohyung},
  booktitle={CVPR},
  year={2022}
}

@inproceedings{wortsman2022robust,
  title={Robust fine-tuning of zero-shot models},
  author={Wortsman, Mitchell and Ilharco, Gabriel and Kim, Jong Wook and Li, Mike and Kornblith, Simon and Roelofs, Rebecca and Lopes, Raphael Gontijo and Hajishirzi, Hannaneh and Farhadi, Ali and Namkoong, Hongseok and others},
  booktitle={CVPR},
  year={2022}
}

@inproceedings{gouk2020distance,
  title={Distance-based regularisation of deep networks for fine-tuning},
  author={Gouk, Henry and Hospedales, Timothy M and Pontil, Massimiliano},
   booktitle={arXiv preprint arXiv:2002.08253},
  year={2020}
}

@inproceedings{izmailov2018averaging,
  title={Averaging weights leads to wider optima and better generalization},
  author={Izmailov, Pavel and Podoprikhin, Dmitrii and Garipov, Timur and Vetrov, Dmitry and Wilson, Andrew Gordon},
  booktitle={UAI},
  year={2018}
}

@inproceedings{aljundi2018memory,
  title="{Memory Aware Synapses: Learning what (not) to forget}",
  author={Aljundi, Rahaf and Babiloni, Francesca and Elhoseiny, Mohamed and Rohrbach, Marcus and Tuytelaars, Tinne},
  booktitle={ECCV},
  year={2018}
}

@inproceedings{zenke2017continual,
  title="{Continual Learning Through Synaptic Intelligence}",
  author={Zenke, Friedemann and Poole, Ben and Ganguli, Surya},
  booktitle={ICML},
  year={2017}
}

@article{kirkpatrick2017overcoming,
  title="{Overcoming Catastrophic Forgetting in Neural Networks}",
  author={Kirkpatrick, James and Pascanu, Razvan and Rabinowitz, Neil and Veness, Joel and Desjardins, Guillaume and Rusu, Andrei A and Milan, Kieran and Quan, John and Ramalho, Tiago and Grabska-Barwinska, Agnieszka and others},
  journal={Proceedings of the national academy of sciences},
  year={2017},
  publisher={National Acad Sciences}
}

@article{van2019three,
  title="{Three Scenarios for Continual Learning}",
  author={van de Ven, Gido M and Tolias, Andreas S},
  journal={arXiv preprint arXiv:1904.07734},
  year={2019}
}

@article{hsu2018re,
  title="{Re-evaluating Continual Learning Scenarios: A Categorization and Case for Strong Baselines}",
  author={Hsu, Yen-Chang and Liu, Yen-Cheng and Ramasamy, Anita and Kira, Zsolt},
  journal={arXiv preprint arXiv:1810.12488},
  year={2018}
}

@inproceedings{rebuffi2017icarl,
  title="{iCaRL: Incremental Classifier and Representation Learning}",
  author={Rebuffi, Sylvestre-Alvise and Kolesnikov, Alexander and Sperl, Georg and Lampert, Christoph H},
  booktitle={CVPR},
  year={2017}
}

@inproceedings{abati2020conditional,
  title="{Conditional Channel Gated Networks for Task-Aware Continual Learning}",
  author={Abati, Davide and Tomczak, Jakub and Blankevoort, Tijmen and Calderara, Simone and Cucchiara, Rita and Bejnordi, Babak Ehteshami},
  booktitle={CVPR},
  year={2020}
}

@inproceedings{liu2021adaptive,
  title="{Adaptive Aggregation Networks for Class-Incremental Learning}",
  author={Liu, Yaoyao and Schiele, Bernt and Sun, Qianru},
  booktitle={CVPR},
  year={2021}
}

@inproceedings{hou2019learning,
  title="{Learning a Unified Classifier Incrementally via Rebalancing}",
  author={Hou, Saihui and Pan, Xinyu and Loy, Chen Change and Wang, Zilei and Lin, Dahua},
  booktitle={CVPR},
  year={2019}
}

@inproceedings{wu2019large,
  title="{Large Scale Incremental Learning}",
  author={Wu, Yue and Chen, Yinpeng and Wang, Lijuan and Ye, Yuancheng and Liu, Zicheng and Guo, Yandong and Fu, Yun},
  booktitle={CVPR},
  year={2019}
}

@inproceedings{romero2014fitnets,
  title="{FitNets: Hints For Thin Deep Nets}",
  author={Romero, Adriana and Ballas, Nicolas and Kahou, Samira Ebrahimi and Chassang, Antoine and Gatta, Carlo and Bengio, Yoshua},
  booktitle = {ICLR},
  year={2015}
}

@inproceedings{zagoruyko2016paying,
	Author = {Zagoruyko, Sergey and Komodakis, Nikos},
	booktitle = {ICLR},
	title = "{Paying More Attention to Attention: Improving the Performance of Convolutional Neural Networks via Attention Transfer}",
	year = {2017}
}

@article{hinton2015distilling,
	Author = {Hinton, Geoffrey and Vinyals, Oriol and Dean, Jeff},
	Journal = {arXiv preprint arXiv:1503.02531},
	title = {{Distilling the Knowledge in a Neural Network}},
	year = {2015}
}

@inproceedings{douillard2020podnet,
  title="{PODNet: Pooled Outputs Distillation for Small-Tasks Incremental Learning}",
  author={Douillard, Arthur and Cord, Matthieu and Ollion, Charles and Robert, Thomas},
  booktitle={ECCV},
  year={2020}
}

@inproceedings{ostapenko2019learning,
  title="{Learning to Remember: A Synaptic Plasticity Driven Framework for Continual Learning}",
  author={Ostapenko, Oleksiy and Puscas, Mihai and Klein, Tassilo and Jahnichen, Patrick and Nabi, Moin},
  booktitle={CVPR},
  year={2019}
}

@inproceedings{shin2017continual,
  title="{Continual Learning with Deep Generative Replay}",
  author={Shin, Hanul and Lee, Jung Kwon and Kim, Jaehong and Kim, Jiwon},
  booktitle={NIPS},
  year={2017}
}

@inproceedings{liu2020generative,
  title="{Generative Feature Replay for Class-Incremental Learning}",
  author={Liu, Xialei and Wu, Chenshen and Menta, Mikel and Herranz, Luis and Raducanu, Bogdan and Bagdanov, Andrew D and Jui, Shangling and de Weijer, Joost van},
  booktitle={CVPR Workshops},
  year={2020}
}

@inproceedings{goodfellow2014generative,
  title="{Generative Adversarial Nets}",
  author={Goodfellow, Ian and Pouget-Abadie, Jean and Mirza, Mehdi and Xu, Bing and Warde-Farley, David and Ozair, Sherjil and Courville, Aaron and Bengio, Yoshua},
  booktitle={NIPS},
  year={2014}
}

@inproceedings{odena2017conditional,
  title="{Conditional Image Synthesis with Auxiliary Classifier Gans}",
  author={Odena, Augustus and Olah, Christopher and Shlens, Jonathon},
  booktitle={ICML},
  year={2017},
}

@inproceedings{rame2022diverse,
  title={Diverse Weight Averaging for Out-of-Distribution Generalization},
  author={Rame, Alexandre and Kirchmeyer, Matthieu and Rahier, Thibaud and Rakotomamonjy, Alain and Cord, Matthieu and others},
  booktitle={NeurIPS},
  year={2022}
}

@inproceedings{wortsman2022model,
  title={Model soups: averaging weights of multiple fine-tuned models improves accuracy without increasing inference time},
  author={Wortsman, Mitchell and Ilharco, Gabriel and Gadre, Samir Ya and Roelofs, Rebecca and Gontijo-Lopes, Raphael and Morcos, Ari S and Namkoong, Hongseok and Farhadi, Ali and Carmon, Yair and Kornblith, Simon and others},
  booktitle={ICML},
  year={2022},
}

@inproceedings{ioffe2015batch,
  title={Batch normalization: Accelerating deep network training by reducing internal covariate shift},
  author={Ioffe, Sergey and Szegedy, Christian},
  booktitle={ICML},
  year={2015}
}

@inproceedings{liu2020mnemonics,
  title="{Mnemonics Training: Multi-Class Incremental Learning without Forgetting}",
  author={Liu, Yaoyao and Su, Yuting and Liu, An-An and Schiele, Bernt and Sun, Qianru},
  booktitle={CVPR},
  year={2020}
}

@inproceedings{kumarFineTuningCanDistort2022,
  title = {Fine-{{Tuning}} Can {{Distort Pretrained Features}} and {{Underperform Out-of-Distribution}}},
  author = {Kumar, Ananya and Raghunathan, Aditi and Jones, Robbie and Ma, Tengyu and Liang, Percy},
  year={2022},
  booktitle = {ICLR},
}

@inproceedings{leeSurgicalFineTuningImproves2022,
  title = {Surgical {{Fine-Tuning Improves Adaptation}} to {{Distribution Shifts}}},
  author = {Lee, Yoonho and Chen, Annie S. and Tajwar, Fahim and Kumar, Ananya and Yao, Huaxiu and Liang, Percy and Finn, Chelsea},
  year = {2023},
  booktitle = {ICLR},
}

@inproceedings{neyshaburWhatBeingTransferred2020,
  title = {What Is Being Transferred in Transfer Learning?},
  booktitle = {NeurIPS},
  author = {Neyshabur, Behnam and Sedghi, Hanie and Zhang, Chiyuan},
  year = {2020},
}

@article{rameModelRatatouilleRecycling2023,
  title={Model Ratatouille: Recycling Diverse Models for Out-of-Distribution Generalization},
  author={Ram{\'e}, Alexandre and Ahuja, Kartik and Zhang, Jianyu and Cord, Matthieu and Bottou, L{\'e}on and Lopez-Paz, David},
  journal={arXiv preprint arXiv:2212.10445},
  year={2022}
}

@inproceedings{stojanovskiMomentumbasedWeightInterpolation2022,
  title = {Momentum-Based {{Weight Interpolation}} of {{Strong Zero-Shot Models}} for {{Continual Learning}}},
  author = {Stojanovski, Zafir and Roth, Karsten and Akata, Zeynep},
  year = {2022},
  booktitle = {NeurIPS Workshops},
}

@inproceedings{tianTrainableProjectedGradient2023,
  title = {Trainable {{Projected Gradient Method}} for {{Robust Fine-tuning}}},
  author = {Tian, Junjiao and Dai, Xiaoliang and Ma, Chih-Yao and He, Zecheng and Liu, Yen-Cheng and Kira, Zsolt},
  year = {2023},
  booktitle = {CVPR},
}

@TechReport{cifar09,
  author= {Krizhevsky, Alex  and  Nair, Vinod and Hinton, Geoffrey},
  title       = "{Learning Multiple Layers of Features from Tiny Images}",
  year        = {2009},
}

@article{ILSVRC15,
Author = {Olga Russakovsky and Jia Deng and Hao Su and Jonathan Krause and Sanjeev Satheesh and Sean Ma and Zhiheng Huang and Andrej Karpathy and Aditya Khosla and Michael Bernstein and Alexander C. Berg and Li Fei-Fei},
Title = {{ImageNet Large Scale Visual Recognition Challenge}},
Year = {2015},
journal   = {IJCV},
}

@inproceedings{wang2022foster,
  title={Foster: Feature boosting and compression for class-incremental learning},
  author={Wang, Fu-Yun and Zhou, Da-Wei and Ye, Han-Jia and Zhan, De-Chuan},
  booktitle={ECCV},
  year={2022},
}

@inproceedings{garipov2018loss,
  title={Loss surfaces, mode connectivity, and fast ensembling of dnns},
  author={Garipov, Timur and Izmailov, Pavel and Podoprikhin, Dmitrii and Vetrov, Dmitry P and Wilson, Andrew G},
  booktitle={NeurIPS},
  year={2018}
}

@inproceedings{zhu2021class,
  title={Class-incremental learning via dual augmentation},
  author={Zhu, Fei and Cheng, Zhen and Zhang, Xu-yao and Liu, Cheng-lin},
  booktitle={NeurIPS},
  year={2021}
}

@inproceedings{cortes2012algorithms,
  title={Algorithms for learning kernels based on centered alignment},
  author={Cortes, Corinna and Mohri, Mehryar and Rostamizadeh, Afshin},
  booktitle={JMLR},
  year={2012},
}

@inproceedings{kornblith2019similarity,
  title={Similarity of neural network representations revisited},
  author={Kornblith, Simon and Norouzi, Mohammad and Lee, Honglak and Hinton, Geoffrey},
  booktitle={ICML},
  year={2019},
}
